\documentclass[12pt]{spieman} 
\usepackage{amsmath,amsfonts,amssymb}
\usepackage{graphicx}
\usepackage{setspace}
\usepackage{tocloft}
\usepackage{textcomp}
\usepackage{color,soul}
\usepackage{subcaption}
\usepackage{float}
\usepackage[colorlinks=true, allcolors=blue]{hyperref}
\usepackage[table,xcdraw,dvipsnames]{xcolor}
\usepackage{lineno}
\usepackage{array}
\usepackage{listings}
\usepackage{mathptmx}
\usepackage{epsfig}
\usepackage{rotating}
\usepackage{makeidx}
\usepackage{url}
\usepackage{booktabs}
\usepackage{tabularx}
\usepackage{blindtext}
\usepackage{times}
\usepackage{nccmath}
\usepackage{mwe}
\usepackage{acro}
\usepackage{adjustbox}
\usepackage{colortbl}
\usepackage{relsize}
\usepackage{pifont}
\usepackage{multirow}
\usepackage{multicol}
\usepackage{nicematrix}
\usepackage{bm}
\usepackage{makecell}
\usepackage{tabu}
\usepackage[capitalize]{cleveref}

\usepackage{color}
\definecolor{dkgreen}{rgb}{0,0.6,0}
\definecolor{gray}{rgb}{0.5,0.5,0.5}
\definecolor{mauve}{rgb}{0.58,0,0.82}

\colorlet{Mycolor1}{green!10!orange}

\title{Robust Fiber Orientation Distribution Function Estimation Using Deep Constrained Spherical Deconvolution for Diffusion MRI}

\author[a]{Tianyuan Yao}
\author[b]{Francois Rheault}
\author[c]{Leon Y Cai}
\author[d]{Vishwesh Nath}
\author[a]{Zuhayr Asad}
\author[a]{Nancy Newlin}
\author[a]{Can Cui}
\author[a]{Ruining Deng}
\author[a]{Karthik Ramadass}
\author[e]{Andrea Shafer}
\author[e]{Susan Resnick}
\author[c]{Kurt Schilling}
\author[a,c,f]{Bennett A. Landman}
\author[a,f,*]{Yuankai Huo}

\affil[a]{Department of Computer Science, Vanderbilt University, Nashville, TN, USA}
\affil[b]{Department of Computer Science, Université de Sherbrooke, Sherbrooke, Québec, Canada}
\affil[c]{Department of Biomedical Engineering, Vanderbilt University, Nashville, TN, USA}
\affil[d]{NVIDIA Corporation, Bethesda, MD, USA}
\affil[e]{Laboratory of Behavioral Neuroscience, National Institute on Aging, Baltimore, MD, USA}
\affil[f]{Department of Electrical and Computer Engineering, Vanderbilt University, Nashville, TN, 37235, USA}

\cftpagenumbersoff{figure}
\cftpagenumbersoff{table} 
\begin{document} 
\begin{sloppypar}
\maketitle
\begin{abstract}
Diffusion-weighted magnetic resonance imaging (DW-MRI) is a critical imaging method for capturing and modeling tissue microarchitecture at a millimeter scale. A common practice to model the measured DW-MRI signal is via fiber orientation distribution function (fODF). This function is the essential first step for the downstream tractography and connectivity analyses. With recent advantages in data sharing, large-scale multi-site DW-MRI datasets are being made available for multi-site studies. However, measurement variabilities (e.g., inter- and intra-site variability, hardware performance, and sequence design) are inevitable during the acquisition of DW-MRI. Most existing model-based methods (e.g., constrained spherical deconvolution (CSD)) and learning based methods (e.g., deep learning (DL)) do not explicitly consider such variabilities in fODF modeling, which consequently leads to inferior performance on multi-site and/or longitudinal diffusion studies. In this paper, we propose a novel data-driven deep constrained spherical deconvolution method to explicitly constrain the scan-rescan variabilities for a more reproducible and robust estimation of brain microstructure from repeated DW-MRI scans. Specifically, the proposed method introduces a new 3D volumetric scanner-invariant regularization scheme during the fODF estimation. We study the Human Connectome Project (HCP) young adults test-retest group as well as the MASiVar dataset (with inter- and intra-site scan/rescan data). The Baltimore Longitudinal Study of Aging (BLSA) dataset is employed for external validation. From the experimental results, the proposed data-driven framework outperforms the existing benchmarks in repeated fODF estimation. The proposed method is assessing the downstream connectivity analysis and shows increased performance in distinguishing subjects with different biomarkers. The plug-and-play design of the proposed approach is potentially applicable to a wider range of data harmonization problems in neuroimaging.

\end{abstract}
{\noindent \footnotesize\textbf{*Corresponding Author:} Yuankai Huo,  \linkable{yuankai.huo@vanderbilt.edu} }

\begin{spacing}{2}  

\section{Introduction}
Diffusion-weighted MRI (DW-MRI) provides a non-invasive approach to estimate the intra-voxel tissue microarchitectures as well as the reconstruction of in-vivo neural pathways of the human brain~\cite{schaefer2000diffusion}. Reproducible fiber orientation distribution function (fODF) estimation is essential for downstream tractography and connectivity analyses~\cite{hagmann2003dti}. Recent advances in imaging technologies, such as high angular resolution diffusion imaging (HARDI)~\cite{tuch2002high}, provide us with a higher angular resolution for modeling intra-voxel orientation uncertainty. On the other hand, diffusion tensor imaging (DTI)~\cite{le2001diffusion}, while useful, does not inherently provide this higher resolution due to its limitations in representing multiple fiber orientations within a single voxel. These new capabilities in imaging, particularly with HARDI, result in more precise depictions of white matter microstructure. However, they also necessitate more sophisticated processing methods due to the increased complexity of the data.

\begin{figure*}[t]
\begin{center}
\includegraphics[width=1\linewidth]{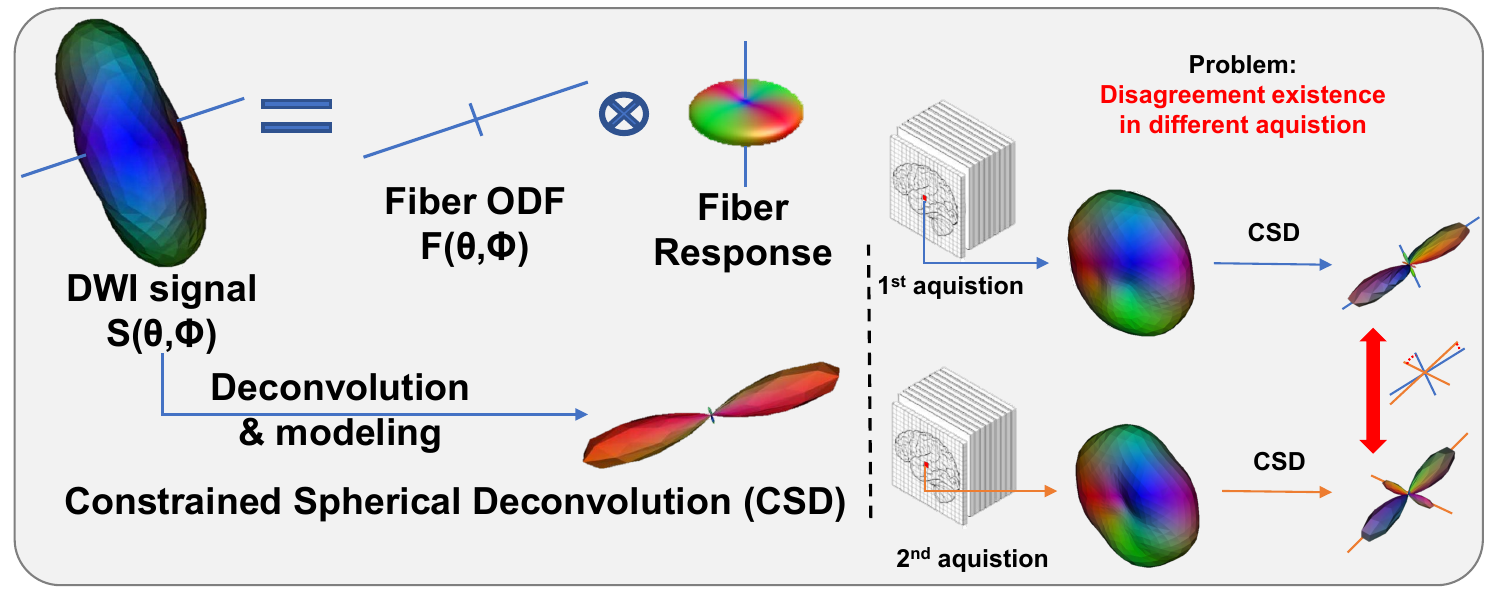}
\end{center}
\caption{\textbf{Overview of the proposed method.} The constrained spherical deconvolution (CSD) is affected by measurement factors of DW-MRI signals (e.g., hardware, reconstruction algorithms, acquisition parameters). The right panel shows the inter-site variability during modeling, even for the data that are collected from the same patient.}
\label{fig:problem}
\end{figure*}

The first family of the ODF esitimation methods is typically called “model-based”, which links underlying tissue microstructures with observed signals via sophisticated mathematical modeling such as constrained spherical deconvolution~\cite{tournier2008resolving,jeurissen2014multi} (CSD), Q-ball~\cite{aganj2010reconstruction}, and persistent angular structure~\cite{jansons2003persistent} (PAS MRI). Among such approaches, CSD is one of the most broadly accepted for modeling HARDI signals [9]. However, CSD is plagued by limited reproducibility (Fig.~\ref{fig:problem}). Several studies have highlighted the biases, inaccuracies, as well as other limitations of HARDI methods in characterizing tissue microstructure (e.g., parameter selection, noise sensitivity, and assumptions)~\cite{schilling2016comparison}. Moreover, such methods exhibit high computational complexity and often require a high number of acquisition points, which might not be available in clinical settings~\cite{benou2019deeptract}.

To address such challenges, the second family of approaches – “data-driven” methods – is attracting increasingly more interest. For example, machine learning (ML) and deep learning (DL) techniques have demonstrated their remarkable abilities in neuroimaging~\cite{poulin2019tractography, benou2019deeptract}. Such approaches have been applied to the task of microstructure estimation, aiming to directly learn the mapping between input DW-MRI scans and output fiber tractography. To maintain the necessary characteristics and reproducibility for clinical translation, robust training is carried out on a diverse and representative dataset in a data-driven method, ensuring the model can handle a variety of patient demographics and scanner variability. The trained model is subsequently validated and tested on independent datasets to evaluate performance and confirm its generalizability. The replicability of the model is assessed, with a thorough evaluation of its consistency across different test sets and scanners. By not assuming a specific diffusion model, data-driven algorithms can reduce the dependence on data acquisition schemes and additionally require less user intervention.

However, measurement variabilities (e.g., inter- and intra-site variability, hardware performance, and sequence design) are inevitable during the imaging process of DW-MRI (Fig.~\ref{fig:problem}). Moreover, most existing model-based and learning-based methods do not explicitly consider such variabilities in modeling, which consequently leads to inferior performance on multi-site and/or longitudinal diffusion studies~\cite{afzali2021sensitivity}. To alleviate such issues, Nath et al.~\cite{nath2019inter} presented a multi-layer perceptron (MLP) based deep learning method for estimating discrete fODF from voxel-wise DW-MRI signals. In this work,  Nath incorporates identical dual networks to minimize the influence of scanner effects via scan-rescan data while learning the mapping between input DW-MRI signals and fiber orientation distribution functions. The novelty in this approach is that paired data is used to drive the training of the network where it learns to ignore features of scanner noise and inter-scanner bias which would otherwise lead the network to differentiate between the data.

Assessing scan-rescan consistency in DW-MRI studies is a common metric for validating the reproducibility of a proposed method because it directly tests the stability of the method under realistic conditions. Variability between multiple scans of the same subject could arise from a number of sources such as slight differences in patient positioning, physiological changes in the patient, or even minor fluctuations in scanner performance. A method that produces consistent results across repeated scans is likely to be more reliable and robust, enhancing its potential for clinical application. We followed this approach to ensure the modeling reproducibility in our study.

However, the 3D context information was largely ignored in the ``voxel-based" MLP model (i.e., model each voxel independently without considering the nearby voxels). It might lead to inferior performance for the 3D DW-MRI signals. To the best of our knowledge, this is the first data-driven approach to explicitly model the 3D patch-wise scan-rescan reproducibility for fODF estimation.

In this paper, we propose a novel deep constrained spherical deconvolution method to explicitly reduce the scan-rescan variabilities, so as to model a more reproducible and robust brain microstructure from repeated DW-MRI scans. Different from the voxel-wise learning in Nath et al.~\cite{nath2019inter}, we introduce a new volumetric patch-based modeling method for 3D DW-MRI signals.  Briefly, the 3×3×3 3D patches from spherical harmonics (SH)-represented DW-MRI signals are employed for a single shell microstructure estimation. The deep convolutional neural network (CNN) is deployed as the computational model in our approach to derive the coefficients.

Another innovation is that we add intra-subject data augmentation in order to alleviate the impacts of a smaller number of diffusion directions on both reproducibility and the accuracy of metrics derived from CSD. The scan/rescan data are employed to facilitate our new loss function in reducing the intra-subject variability. The method has been trained, validated, and tested on both the HCP young adults (HCP-ya) test-retest group~\cite{van2013wu} and the MASiVar dataset~\cite{cai2021masivar}. To assess the model’s generalizability, we applied our model to the Baltimore Longitudinal Study of Aging (BLSA) dataset as the external validation~\cite{ferrucci2008baltimore}. With both direct deployment and further finetuning, we witnessed an increasing consistency between intra-subject scans. Additionally, the brain structural connectomes are computed from the deep CSD as the downstream task for further model evaluation.

Our contribution is four-fold:
\begin{itemize}

\item We propose a novel deep constrained spherical deconvolution method to explicitly reduce the scan-rescan variabilities, so as to model a more reproducible and robust brain microstructure from repeated DW-MRI scans. To the best of our knowledge, this is the first data-driven approach to explicitly model the 3D patch wise scan-rescan reproducibility for fODF estimation.

\item Different from the previous voxel-wise learning studies, we introduce a new 3D volumetric representation of DW-MRI signals for a single-shell microstructure estimation.

\item We propose a new intra-subject augmentation strategy that increases model robustness under the “fewer diffusion directions” scenarios.

\item The proposed method is a “plug-and-play” design as a simple multi-layer regression network, which can be easily aggregated with downstream connectivity analysis.

\end{itemize}

\section{Related work}

Ensuring reproducibility has been an important research topic in MR imaging, specifically in diffusion imaging~\cite{fortin2017harmonization, pinto2020harmonization}. To control the site effect in multi-site imaging data, several strategies have been proposed. Such approaches can be summarized into two major categories: conventional statistics-based method methods~\cite{rao2017predictive, fortin2018harmonization, fortin2017harmonization, pomponio2020harmonization, xia2019reproducibility,yu2018statistical} and more recent machine learning based methods~\cite{dewey2019deepharmony, zuo2021unsupervised, zhao2019harmonization}.

\subsection{Statistics-based Methods}
Conventional statistical methods are usually applied in a linear regression manner on univariate metrics with sites indexed as a categorical covariate, such as the least squares-based general linear model~\cite{rao2017predictive} and Bayesian estimation-based ComBat~\cite{fortin2018harmonization, fortin2017harmonization}. These methods have been utilized in multisite imaging studies and have shown a powerful capacity for removing linear site effects in brain metrics~\cite{pomponio2020harmonization, xia2019reproducibility,yu2018statistical}. However, noticeable limitations have been observed for this type of method. First, the site effect is mathematically assumed to be linear, while the actual effect can be fundamentally more complicated. Second, brain characteristics are considered independently in these models, largely neglecting the spatial and topological relationships among brain regions.

\subsubsection{Machine Learning based Approaches}
Recently proposed DL-based harmonization methods, including U-Net~\cite{dewey2019deepharmony}, cycle generative adversarial network~\cite{modanwal2020mri}, or three-dimensional convolutional neural network ~\cite{tong2020deep}, allow for mapping the complex abstract representations of the nonlinear spatial pattern of the site effects. These models have been primarily applied to ensure reproducibility of diffusion images~\cite{moyer2020scanner}, structural images ~\cite{zuo2021unsupervised}, and morphological measurements~\cite{zhao2019harmonization}, successfully eliminating the site effect with complex spatial or topological information. The model training strategy of site pairing is a common approach for DL-based methods. Briefly, the fusion of data from multiple sites from a single model greatly increases the generalizability of deep learning models. Most methods are ensuring test-retest reliability on the signal level.

As for tractography, most DL-based methods are dealing with the modeling process~\cite{karimi2021learning, lin2019fast, nath2020deep}. By regarding different data as the gold standard, these approaches are chasing perfect fitting with different ground truth (GT) data, but the feasibility/confidence of the GT remains risky. For instance, Nath et al~\cite{nath2019inter} used monkey histology data as GT for fiber orientation distributions. Such data is hard to obtain for a large human cohort, risking the generalizability of the trained model. Sedlar et al.~\cite{sedlar2021diffusion} regarded the human data from HCP as GT, which has over three shells with b-values of 1000, 2000, and 3000 $s/mm^{2}$ (each with 90 gradient directions). However, such acquisition settings are not common in clinical settings. Additionally, even the same subject collected at two different settings may have some degrees apart.  As a result, they might have very similar diffusion signals but fundamentally different fODFs. In our study, we proposed a robust fitting with GT while constraining the scan-rescan variabilities through modeling.

\section{Method}

\subsection{Data Representation}
Spherical Harmonics (SH) are functions defined on the sphere. A collection of SH can be used as a basis function to represent and reconstruct any function on the surface of a unit sphere~\cite{garyfallidis2014dipy}.  All diffusion signals are transformed to SH basis signal ODF as a unified input for deep learning models, using spherical harmonics with the ’tournier07’ basis~\cite{tournier2019mrtrix3}. For the spherical harmonics coefficients $c^m_k$, $k$ is the order, $m$ is the degree. For a given value of $k$, there are $2m + 1$ independent solutions of this form, one for each integer $m$ with $-k \le m \le k$. In practice, a maximum order $L$ is used to truncate the SH series. By only taking into account even-order SH functions, the above bases can be used to reconstruct symmetric spherical functions.

\subsection{Architecture}
\begin{figure*}[t]
\begin{center}
\includegraphics[width=1\linewidth]{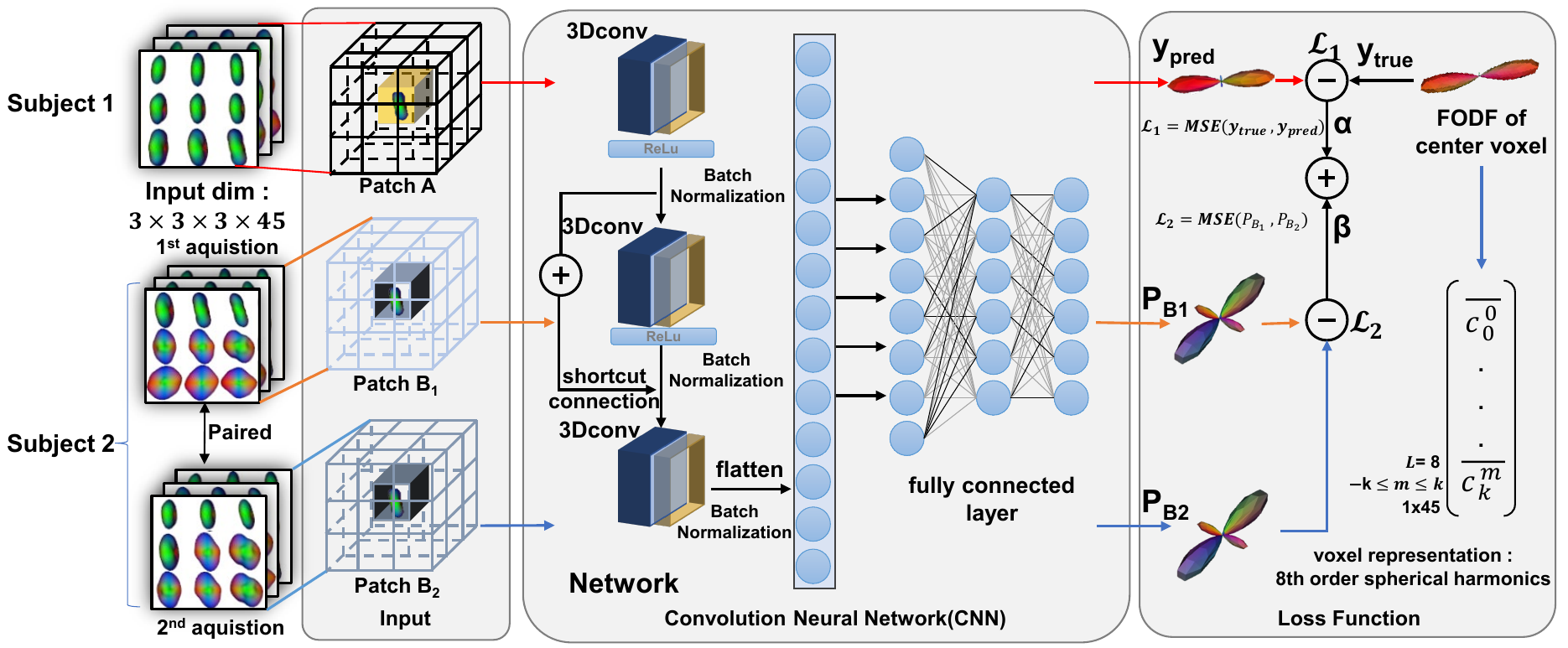}
\end{center}
\caption{\textbf{Deep learning architecture.} A 3D patch-wise convolutional neural network is proposed to fit the fiber orientation distribution function (fODF) from the spherical harmonics (SH) using $3\times3\times3$ cubic DW-MRI signals. A contrastive loss is introduced to reduce the intra-subject variability.}
\label{fig:model}
\end{figure*}

Inspired by Nath et al.~\cite{nath2019inter,nath2020deep}, we employ a 3D CNN with a residual block and utilize $3\times3\times3$ cubic patches as inputs. The rationale is that 3D patches might provide more complete spatial information for deep learning networks (Fig.~\ref{fig:model}) compared with modeling each individual voxel independently. $8^{th}$ order SH is used as data representation in our study. Briefly, the input size of the network is $3\times3\times3$ with 45 channels, while the outputs are 45 $8^{th}$ order spherical harmonics coefficients. During training, the architecture takes 3 patches as input, where the first comes from one subject, the network learns the direct mapping between DW-MRI signals and fiber orientation distribution functions of the center voxel. The other two are paired patches extracted from scan/rescan DW-MRI of another subject and the network learns to minimize the difference of the fiber orientation distribution functions of the center voxel. For validation of the proposed method, it is in terms of accuracy relative to withheld voxels and reproducibility with paired voxels in scan-rescan imaging.

In this network, three subsequent convolutional layers serve as the critical components of the CNN with 3D convolutional filters. One residual block is included to allow for the direct transmission of information from input and the third convolutional layer, thereby enabling the effective training of deep learning networks. The convolutional filters are then flattened and connected to two dense layers for predicting fODF at the center voxel locations.  All layers have ReLU as the activation function. To enhance model training efficiency and stability, we employ batch normalization (BN) in our deep learning architecture to standardize the inputs to each layer of the network, reducing internal covariate shifts.

\subsection{Loss Function}
We introduce a customized loss function, as shown in~\cref{loss,loss2,loss3}. The first term is the MSE loss between the network’s FOD prediction and the ground truth with the hyper-parameter ‘$\alpha$’. N is the number of samples, $m$ is the order of the spherical harmonic (SH) basis and $c^m_k$ is the SH coefficients. The second term is the MSE loss between a corresponding/paired of voxels ($u$ and $v$). The second term has an expectation to be 0 and the hyper-parameter is ‘$\beta$’. Specifically, if no scan/rescan data participate during training, ‘$\beta$’ is set to 0.

\begin{equation}\label{loss}
\begin{split}
 	loss1=\frac{1}{N}\sum_{i=1}^{N}\sum_{k=0}^{L}\sum_{m=-k}^{k}((c^m_k)_{true,i}-(c^m_k)_{pred,i})^2\\
\end{split}
\end{equation}

\begin{equation}\label{loss2}
\begin{split}
	loss2 = \frac{1}{N}\sum_{i=1}^{N}\sum_{k=0}^{L}\sum_{m=-k}^{k}((c^m_k)_{u,i}-(c^m_k)_{v,i})^2\\
\end{split}
\end{equation}

\begin{equation}\label{loss3}
\begin{split}
	Loss = \alpha*loss1 + \beta*loss2 \\  
\end{split}
\end{equation}

\subsection{Intra-subject data augmentation}
In order to provide a robust microstructure estimation, we introduce intra-subject data augmentation during our network training. By performing random diffusion directions dropout and feeding the model with augmented data during training, the model learns to handle situations like missing or corrupt diffusion directions due to factors like patient movement or hardware malfunction and make accurate predictions even with incomplete data. Furthermore, this augmentation strategy can improve the model's generalizability by exposing it to a wider range of data scenarios, thus enabling it to better handle the variability and complexity inherent in real-world diffusion data. An additional b-vector check is performed to ensure the rest directions are still well distributed on a sphere. Thus, we have reconstruction results from different total numbers of diffusion directions from the same DW-MRI signal. The CSD methods are sensitive to the number of diffusion directions and therefore the generated fODF are augmented. By applying this augmentation, the diffusion signal ODF generated from fewer diffusion directions is labeled by the CSD results with the full numbers of gradient directions during the training process.

\section{Experiment}
The experiments can be summarized as with/without scan/rescan data, and with/without intra-subject augmentation on two deep learning models (voxel-wise MLP presented by Nath et al.~\cite{nath2019inter} and ours in Fig.~\ref{fig:model}). We assessed the models, as well as benchmarks, using the overall mean ACC on white matter voxels between the prediction and the ground truth on the HCP and MASiVar dataset. The BLSA dataset has been introduced to further test model reproducibility.

We conducted studies to examine the robustness of the model with ‘fewer gradient directions.’ The results are quantitatively evaluated with single shell single tissue CSD (ssst-CSD). Full direction ssst-CSD is regarded as the silver standard.

Eventually, complex network measures of brain structural connectomes (modularity, average betweenness centrality, characteristic path length, and global efficiency) are computed as an example of the downstream task evaluation~\cite{rubinov2010complex}.

\subsection{Data \& Data process}
For the HCP-ya dataset~\cite{van2013wu}, 45 subjects with the scan-rescan acquisition were used (a total of 90 images). A T1 volume of the same subject was used for WM segmentation using SLANT~\cite{huo20193d}. HCP was distortion corrected with topup and eddy~\cite{andersson2003correct, andersson2016integrated}. The acquisitions at b-value of 1000 and 2000 $s/mm^{2}$ (each shell with 90 diffusion directions) were extracted for the study. 30 subjects were used as training data, 10 were used for validation, and 5 were used for testing.

For the MASiVar dataset~\cite{cai2021masivar}, five subjects were acquired on three different sites referred to as ‘A’, ‘B’ and ’C’. Structural T1 was acquired for all subjects at all sites. All in-vivo acquisitions were pre-processed with the PreQual pipeline~\cite{cai2021prequal} and then registered pairwise per subject. The acquisitions at b-value of 1000 and 2000 $s/mm^{2}$ (each shell with 96 diffusion directions) were extracted for the study. Two subjects from sets ’A’ and ‘B’ were used as paired training data. One subject was used for validation, while two subjects were used for testing. Scans from site ‘C’ were only used for evaluation.

Twenty-four subjects with the scan-rescan acquisition from The Baltimore Longitudinal Study of Aging (BLSA)~\cite{ferrucci2008baltimore} were used as an additional model evaluation cohort. The BLSA dataset was acquired at a b-value of 700 $s/mm^{2}$ using a Philips 3T scanner. The data were pre-processed with Pre-Qual. 18 subjects were used for training, 2 subjects were for validation, and 4 subjects were for testing. Additionally, another cohort of 198 subjects with biological information (Age, Sex, and $\epsilon4$ allele of apolipoprotein E (APOE) states) was used for analyzing connectomics. Between them, 46 subjects were APOE positive while 152 subjects were APOE negative. They were pre-processed with the same protocol.

\subsection{Model-based methods}
We processed all the data to generate fODF with single shell single tissue CSD (ssst-CSD) using the DIPY library (version 1.50) with its default setting~\cite{garyfallidis2014dipy}. Reconstructions from full diffusion direction DMRI were regarded as ground truth. Voxel-wise agreement (metrics see \ref{metrics}) between paired data was calculated as silver standard.

\subsection{Machine learning-based methods}
HCP and MASiVar datasets were employed to evaluate the learning-based method. The image data at a b-value of 2000 $s/mm^{2}$  were used. We use a voxel-wise neural network as our deep learning baseline. The network consists of four fully connected layers. The number of neurons per layer is 400, 45, 200, and 45. The input is the $1\times45$ vector of the SH basis signal ODF, and the output is the $1\times45$ vector of the SH basis Fiber ODF. The architecture in Fig.~\ref{fig:model} was used to evaluate the proposed patch-wise deep learning experiment.

Furthermore, the generalizability of different approaches was assessed by a rigorous external validation using BLSA dataset. The BLSA dataset was acquired at a b-value of 700 $s/mm^{2}$. We used the image data at b-value of 1000  $s/mm^{2}$ from both HCP and MASiVar to train the deep learning model using the same approaches in the previous experiments. Briefly, we trained the deep learning model with only BLSA data as baseline performance. To further test our model capacity, we further assess the scenarios of finetuning the last two linear layers, beyond applying our trained model on BLSA. Thus, we have three general approaches to compare the performance. Moreover, with/without scan/rescan data, and with/without intra-subject augmentation on two deep learning models (voxel-wise MLP presented by Nath et al.\cite{nath2019inter} and our model in Fig.~\ref{fig:model}) are included as additional evaluation.

\subsection{Ablation study}
In the ablation study, we evaluated the intra-subject consistency on all white matter voxels with a different number of diffusion directions. The deep learning model which had the best performance on the validation set is chosen for comparison. Different reconstruction results from a different number of diffusion directions were visualized as the qualitative results, while their agreements with the silver standard were assessed as the quantitative results.

Additionally, we evaluated the effectiveness of the random dropout of diffusion directions using the MASiVar dataset. A subset of 45 diffusion directions from the same shell was first determined. A number of 45 was the basic requirement for $8^th$ order spherical harmonics and then visualized. Then the dropout (drops from 96 to the subset) was performed randomly 10 times. With both results of CSD reconstruction and DL prediction on the DW-MRI signal, the ACC and the MD (0th order SH of the fODF) were computed. The spheres in both single fiber and crossing fiber areas were presented to assess the effectiveness of the intra-subject augmentation by examining shapeshift.

\subsection{Downstream task evaluation}
To evaluate the performance on downstream tasks, we predicted APOE status using different brain structure connectome maps via different fODF modeling methods. Complex network measures of brain structural connectomes (modularity, average betweenness centrality, characteristic path length, and global efficiency) were computed from the results as the downstream task~\cite{rubinov2010complex}. During the process, fODFs were computed from both CSD and our DL networks. We used the MRTrix (version-3.0.3) default probabilistic tracking algorithm of second-order integration over fiber orientation distributions (FODs) for tractography~\cite{tournier2010improved}. We generated 10 million streamlines to build each tractogram, seeding and termination using the five-tissue-type mask. We allowed backtracking. After, we converted the tractography to a connectome with the Desikan-Killany atlas~\cite{klein2012101} with 84 cortical parcellations from Freesurfer~\cite{fischl2012freesurfer}. Graph theory measures were computed with the Brain Connectivity Toolbox (BCT, version-2019-03-03)~\cite{rubinov2010complex}.

Modularity was the degree to which the network may be subdivided into clearly delineated and nonoverlapping groups~\cite{rubinov2010complex}. Betweenness centrality was the fraction of the shortest paths in the network that contained a given node~\cite{rubinov2010complex}. Average betweenness centrality was the average fraction of shortest paths that nodes in a network participate in~\cite{rubinov2010complex}. The characteristic path length was the average shortest path between nodes in millimeters. Global efficiency was the average inverse shortest path length~\cite{rubinov2010complex}. Using the graph measures and biological information (age, sex) as input, we used a three-layer MLP network to perform classification across different APOE groups between CSD and the deep learning approaches.

We performed leave-one-out cross-validation as the evaluation strategy. Weighted cross-entropy was applied as we had unbalanced groups. The cross-validation had been performed 20 times, and the 95

\subsection{Evaluation metrics}\label{metrics}
To compare the predictions of the proposed deep learning methods, we used angular correlation coefficient (ACC, Eq.~\ref{ACC}) to evaluate the similarity of the prediction when compared with the ground truth estimate of CSD. ACC was a generalized measure for all fiber population scenarios. It assessed the correlation of all directions over a spherical harmonic expansion. It is calculated between fODF of two voxels ($u$ and $v$), where $u_{km}$ and $v_{km}$ are the SH coefficients. In brief, it provided an estimate of how closely a pair of fODF’s were related on a scale of -1 to 1, where 1 was the best measure. Here ‘u’ and ‘v’ represented sets of SH coefficients.

\begin{equation}\label{ACC}
\begin{split}
ACC= \frac{\sum_{k=1}^{L}\sum_{m=-k}^{k}(u_{km})(v^*_{km})}{[\sum_{k=1}^{L}\sum_{m=-k}^{k}|u_{km}|^2]^{0.5}\cdot[\sum_{k=1}^{L}\sum_{m=-k}^{k}|v_{km}|^2]^{0.5}} 
\end{split}
\end{equation}

For downstream task evaluation, we used the classical classification metrics (Accuracy, precision, recall, and F1) to evaluate the biomarker prediction. Macro precision, recall, and F1 were used in our study as we had unbalanced positive/negative APOE groups.

\section{Results}

\subsection{Deep learning results}
The results of the fODF estimation are presented in Table~\ref{table:performance1}. The mean ACC over white matter regions are shown in the table to evaluate the similarity of the prediction when compared with the truth estimate of CSD.  This serves as the accuracy of prediction. The mean ACC over white matter regions are calculated in the test cohort of scan/rescan imaging, shown as scan/rescan consistency in the table. This serves as validation of reproducibility.

The implementation of the CNN network for 3D-patch inputs led to a superior spherical harmonics coefficients estimation by incorporating more information from neighboring voxels. Meanwhile, by introducing the identity loss with scan/rescan data, the proposed method achieved a higher consistency while maintaining higher angular correlation coefficients with CSD.

\begin{table}[t]
\centering
\caption{Performance of fODF prediction on HCP\&MASiVar}
\begin{tabular}{ccccc}
\toprule
model \& Method                & scan/rescan & \makecell{intra-subject \\ augmentation}   & \makecell{ACC, compare with \\  full direction CSD}   & \makecell{scan/rescan \\ consistency}  
 \\
\midrule
CSD (silver standard)            & N/A        & N/A    &   1         & 0.826 \\
\midrule
\multirow{3}{*}{voxel-wise}  &      &         & 0.942     & 0.830 $\textcolor{red}{\bm{\star}}$   \\
                              & \checkmark       &    & 0.938     &0.878 $\textcolor{red}{\bm{\star}}$      \\
                              & \checkmark      & \checkmark     & 0.939      & 0.882 $\textcolor{red}{\bm{\star}}$ \\

\midrule
\multirow{3}{*}{patch-wise}     &     &     & 0.949    &  0.834 $\textcolor{red}{\bm{\star}}$  \\
                              & \checkmark   &       &  \textbf{0.954}     &  0.886 $\textcolor{red}{\bm{\star}}$  \\
                              & \checkmark   & \checkmark       & 0.953        & \textbf{0.891} $\textcolor{red}{\bm{\star}}$\\

\bottomrule
\end{tabular}
\text{*Mean ACC are calculated over white matter voxels.}
\newline
\label{table:performance1}
\caption*{Table 1: Mean ACC over white matter region voxels between 1) prediction and ground truth signals, 2) prediction of paired scan/rescan DWI signals are calculated. The Wilcoxon signed-rank test is applied to voxel-wise ACC as a statistical assessment between the derived results with silver standard CSD. The statistical difference ($p < 0.001$) compared with scan/rescan consistency from CSD is marked as $\textcolor{red}{\bm{\star}}$.}
\end{table}

\subsection{Model evaluation on unseen dataset}
By applying our method on BLSA dataset, as shown in Table~\ref{table:performance3}. It shows a great improvement in scan/rescan consistency while applying our approaches and using BLSA as finetuning data as compared with both the silver standard and the model trained directly on the BLSA dataset (0.838 vs. 0.834 vs. 0.635). More importantly, by directly applying the model to unseen data, we still show significant intra-subject consistency and maintain high agreement (0.836 vs. 0.872) with the full direction CSD.

\begin{table}[htbp]
\caption{Performance on BLSA dataset}
\begin{adjustbox}{width=\textwidth}
\begin{NiceTabular}{ccccccc}
\toprule
model \& method   &training data &test data   & scan/rescan  & \makecell{intra-subject \\ augmentation}   & \makecell{ACC, compare with \\  full direction CSD}  & \makecell{scan/rescan \\ consistency}  
 \\
\midrule
CSD (silver standard)   & N/A  & BLSA & N/A  & N/A  & 1 (upper bound)     & 0.635 \\ 
\midrule

\multirow{2}{*}{\makecell{voxel-wise DL \\ (Nath et al. 2019)}}  &\cellcolor[gray]{0.9} HCP, MASiVar &\cellcolor[gray]{0.9} BLSA  &\cellcolor[gray]{0.9}   &\cellcolor[gray]{0.9}   &\cellcolor[gray]{0.9} 0.845 & \cellcolor[gray]{0.9}0.747 $\textcolor{red}{\bm{\star}}$\\
& HCP, MASiVar & BLSA  &  \checkmark   &    & 0.832  &  0.813 $\textcolor{red}{\bm{\star}}$\\
\midrule

\multirow{6}{*}{Patch-wise DL}  &\cellcolor[gray]{0.9} HCP, MASiVar &\cellcolor[gray]{0.9} BLSA  & \cellcolor[gray]{0.9} &\cellcolor[gray]{0.9}  & \cellcolor[gray]{0.9}0.829  &\cellcolor[gray]{0.9} 0.763 $\textcolor{red}{\bm{\star}}$\\

& HCP, MASiVar & BLSA   &   \checkmark  & &0.834   &0.812 $\textcolor{red}{\bm{\star}}$  \\

& \cellcolor[gray]{0.9}HCP, MASiVar & \cellcolor[gray]{0.9}BLSA     & \cellcolor[gray]{0.9} \checkmark  & \cellcolor[gray]{0.9}\checkmark  &\cellcolor[gray]{0.9}0.836 &\cellcolor[gray]{0.9}0.824 $\textcolor{red}{\bm{\star}}$\\
\cline{2-7}
& BLSA & BLSA     &  \checkmark    &  &\textbf{0.872}    &0.834 $\textcolor{red}{\bm{\star}}$\\
\cline{2-7}
& \makecell{\cellcolor[gray]{0.9}HCP, MASiVar, \\ \cellcolor[gray]{0.9}BLSA (finetune)} & \cellcolor[gray]{0.9}BLSA &\cellcolor[gray]{0.9}\checkmark &\cellcolor[gray]{0.9}       &\cellcolor[gray]{0.9}0.849    &\cellcolor[gray]{0.9}0.828 $\textcolor{red}{\bm{\star}}$\\

& \makecell{HCP, MASiVar, \\ BLSA (finetune)} & BLSA &\checkmark & \checkmark &0.842   &\textbf{0.838} $\textcolor{red}{\bm{\star}}$\\

\bottomrule
\end{NiceTabular}
\end{adjustbox}
\newline
\caption*{Table 2: The performance of different methods on the BLSA dataset. Wilcoxon signed-rank test is applied to compare the derived results with silver standard CSD. The statistical difference ($p < 0.001$) compared with scan/rescan consistency from CSD is marked as $\textcolor{red}{\bm{\star}}$. }
\label{table:performance3}
\end{table}

\subsection{Ablation study}
\begin{figure*}[htbp]
\begin{center}
\includegraphics[width=1\linewidth]{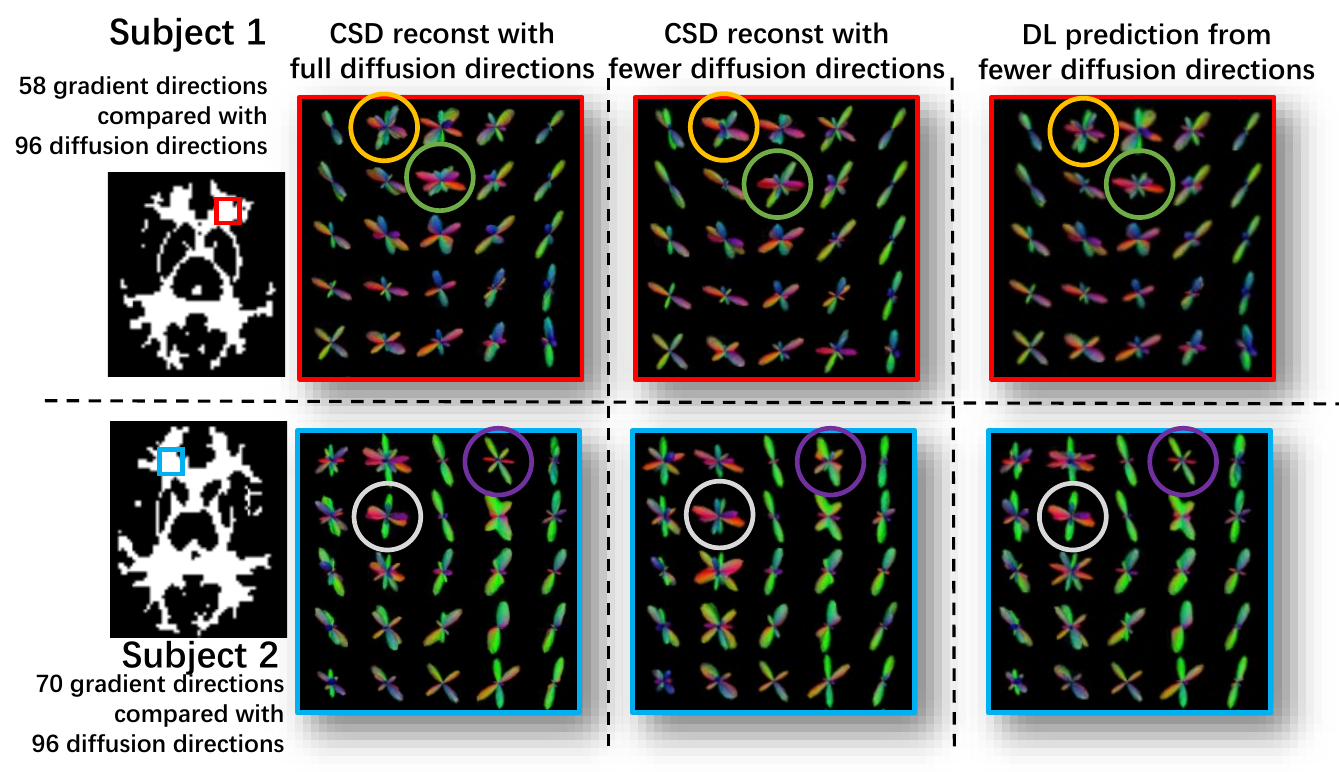}
\end{center}
\caption{\textbf{Qualitative results of fODF modeling.} Visualizations of fODF of the proposed deep learning (DL) method and the results from CSD modeling on two testing subjects in MASiVar. We took the same patch(matched with same color of the border) from the results in the crossing fiber area. The same voxel is matched with the same color of the circle.}
\label{qualitative}
\end{figure*}

\begin{figure*}[htbp]
\begin{center}
\includegraphics[width=1\linewidth]{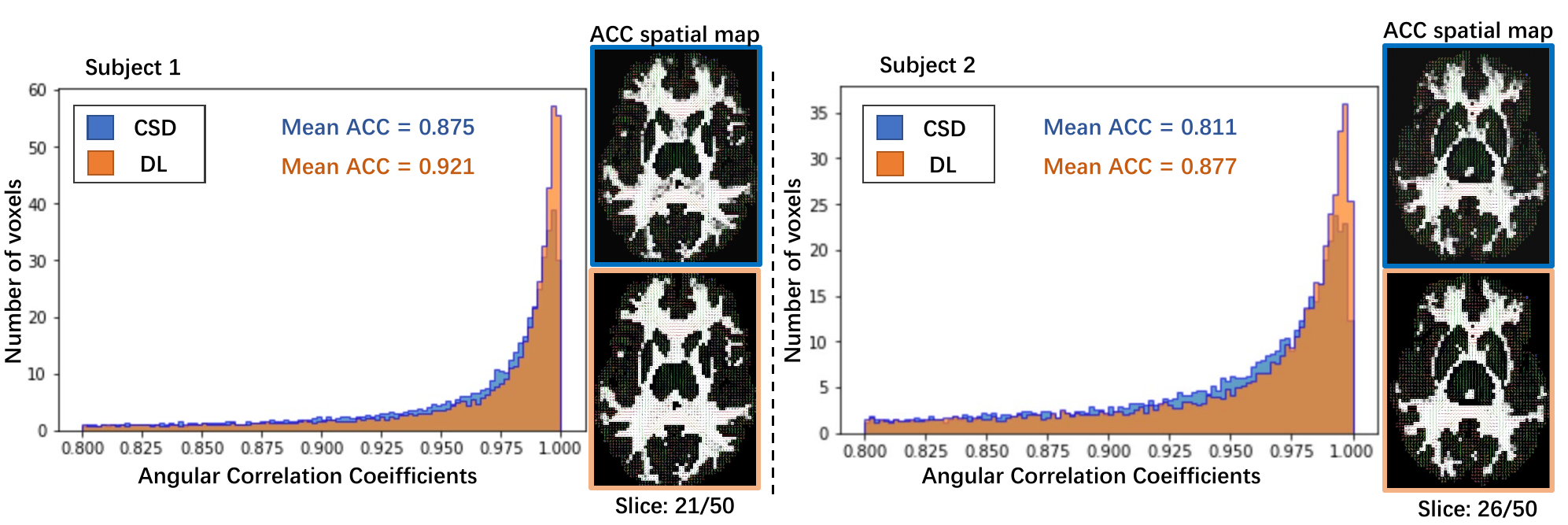}
\end{center}
\caption{\textbf{ACC histogram and ACC spatial map.} This figure depicts the histogram of ACC between full diffusion directions' reconstruction and fewer directions' reconstruction while using the proposed deep learning (DL) method and the results from CSD modeling on the 2 testing subjects in MASiVar. The ACC spatial maps are the comparison between (1) the fODFs of reconstruction from CSD with full diffusion directions and (2) fewer diffusion directions' CSD and DL estimator on two testing subjects.}
\label{histogram}
\end{figure*}

In the ablation study (Table~\ref{table:performance2}), we evaluate the intra-subject augmentation by comparing the intra-subject consistency on all white matter voxels with different number of diffusion directions. The deep learning model which has the best performance on the validation set is chosen for comparison. In Fig.~\ref{qualitative}, the right side shows a qualitative result of the visualization of the estimated SH coefficients and the left side shows the comparison with full-direction CSD. By performing CSD, the test subjects with a mean of 72 diffusion directions can only maintain a mean ACC of 0.848 as compared with their same acquisition with 96 directions. By adding the intra-subject augmentation during the training process, both voxel-wise and patch-wise models have significant improvement, which shows that deep learning reveals untapped information during the ODF estimation. Fig.~\ref{histogram} shows the result of 1) estimation on a signal with fewer diffusion directions using a patch-wise DL model with scan/rescan data and intra-subject augmentation participated during training and 2) CSD reconstruction result.

\begin{figure*}[htbp]
\begin{center}
\includegraphics[width=1\linewidth]{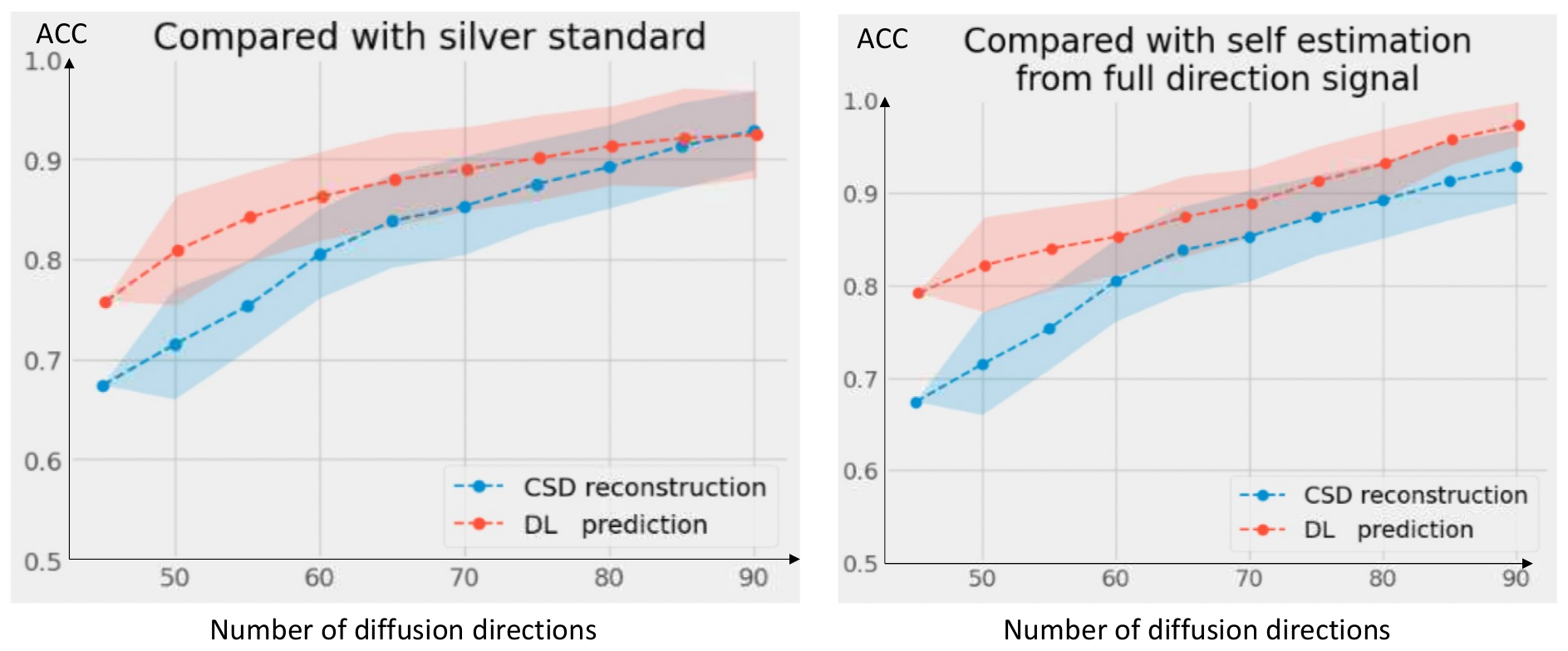}
\end{center}
\caption{\textbf{Quantitative result of performances of methods/modeling with diffusion direction dropout.} ACC is calculated at specific intervals - every 5 diffusion directions - such as at 45, 50, 55, and so on, enabling us to evaluate how the consistency of the model's/method's output was preserved despite the reduction in diffusion gradient directions. The dropout (drops from 96 to the subset of 45 directions) was performed randomly 10 times. The mean ACC and the std are calculated and shown in the line chart. The left panel shows that the deep learning-based method maintains high consistency than the CSD reconstruction when both are compared with the silver standard (full-direction CSD) during the diffusion direction dropout. The right panel presents a similar assessment, except comparing itself using the full-direction modeling.}
\label{degradation}
\end{figure*}

We're evaluating our deep learning model by testing its performance with degraded signals, which involve artificially reducing input data quality (diffusion direction dropout). This assessment allows us to understand the model's robustness under less ideal conditions by comparing its results with CSD. Fig.~\ref{degradation} shows how different methods maintain consistency in their result compared with silver standard and methods' self-full direction-estimation when faced with degraded input data.

\begin{figure*}[htbp]
\begin{center}
\includegraphics[width=1\linewidth]{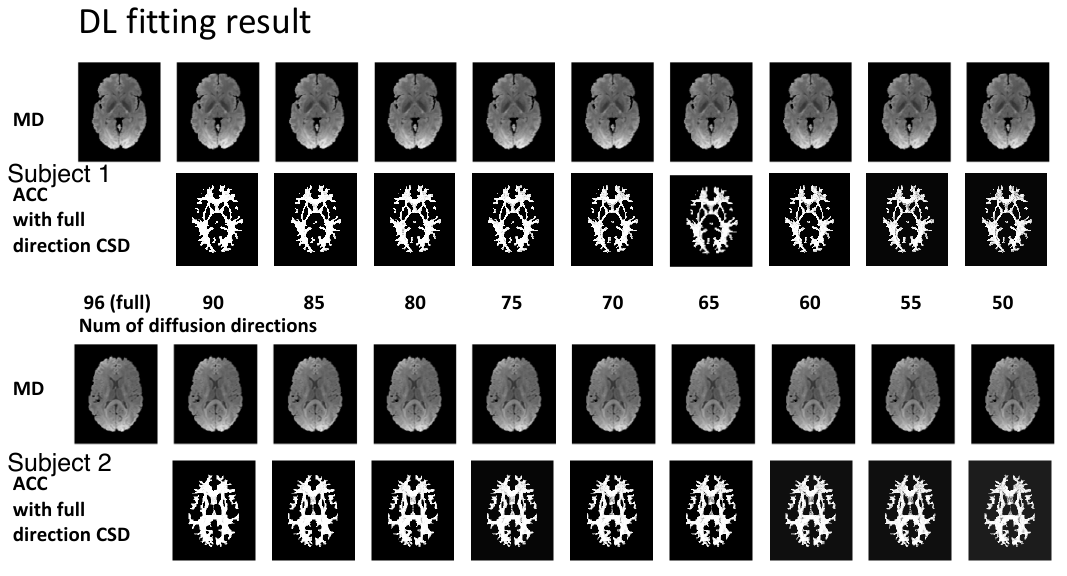}
\end{center}
\caption{\textbf{Qualitative visualization of the predicted coefficients in fewer diffusion directions scenario.} The mean diffusivity map(MD, $0^{th}$ order of the SH coefficient)  and ACC agreement spatial map (even orders of the SH coefficient without $0^{th}$ order,  compared with the silver standard--full direction CSD) of deep learning results of the two MASiVar test subjects. The dropout was performed from 96 to the subset of 45 directions while the visualization was shown at intervals of every 5 diffusion directions.}
\label{MD}
\end{figure*}

\begin{figure*}[htbp]
\begin{center}
\includegraphics[width=0.9\linewidth]{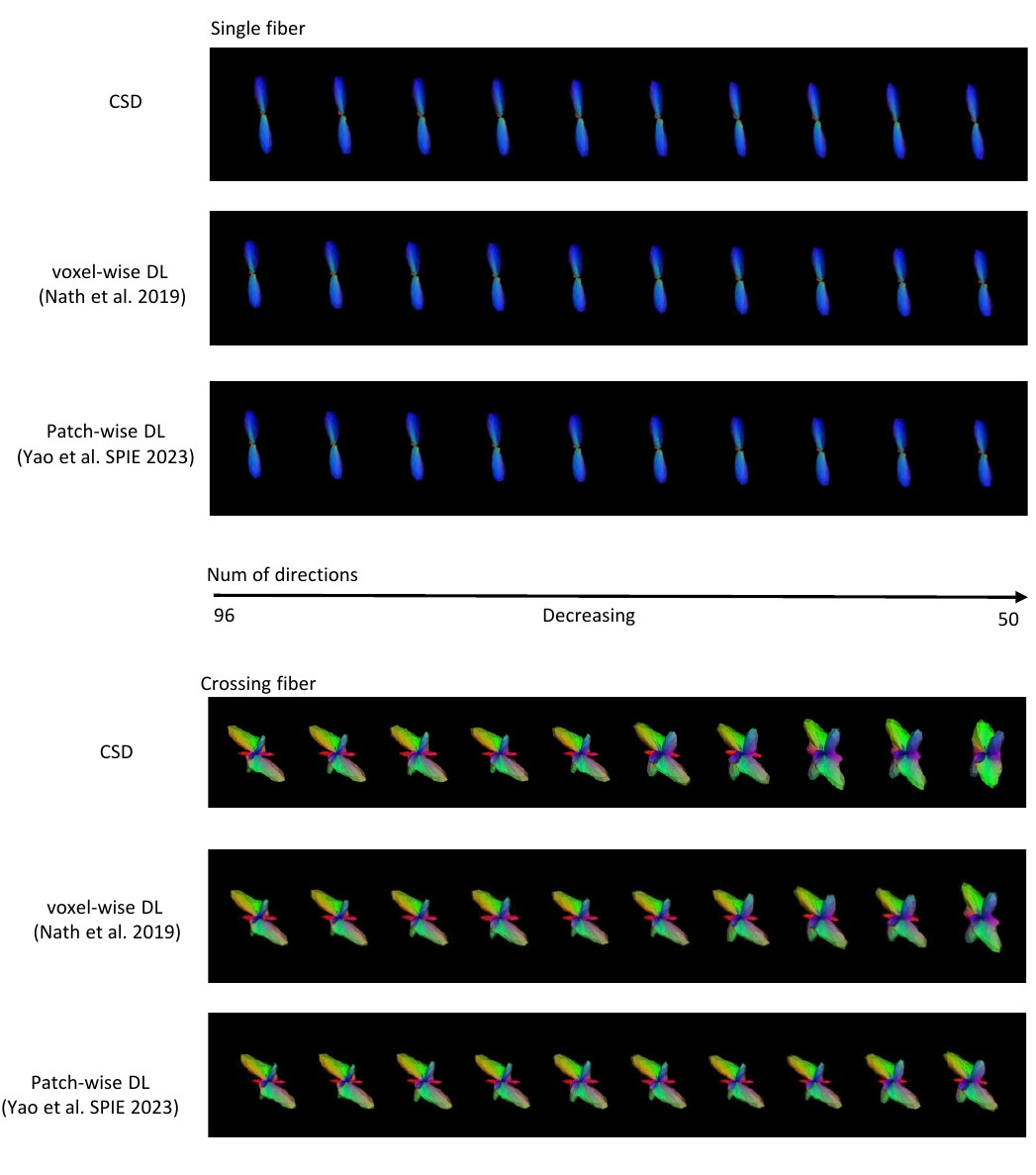}
\end{center}
\caption{\textbf{Qualitative results of synthetic spheres.} Visualization of reconstruction from different numbers of diffusion directions of voxel from single/crossing fiber population. The dropout was performed from 96 to the subset of 45 directions while the visualization was shown at intervals of every 5 diffusion directions. CSD reconstruction of a voxel in crossing fiber population has an obvious shapeshift in the low-resolution scheme while deep learning with the data augmentation strategies remains a robust estimation.}
\label{voxel vis}
\end{figure*}

Fig.~\ref{MD} and Fig.~\ref{voxel vis} provide an additional qualitative visualization of the robust deep learning fitting during diffusion direction dropout. In Fig.~\ref{MD} we plot the mean diffusivity and the ACC spatial map where MD indicates the fitting of the $0^{th}$ order coefficients (No.1 in 45) and the ACC focus on the rest of the SH coefficients (No.2-45 in 45). In Fig.~\ref{voxel vis}, we focus on two unique voxels belonging to the single fiber population and crossing fiber population respectively, the result shows that CSD reconstruction has an obvious shapeshift in the low-resolution scheme while deep learning with the data augmentation strategies remains higher consistency, especially in the visualization of the crossing-fiber voxel.

\begin{table}[h]
\centering{}
\caption{Performance in fewer diffusion direction situation}
\begin{tabular}{cccc}
\toprule
model \& method   & scan/rescan & \makecell{intra-subject \\ augmentation}   & \makecell{intra-subject \\ consistency}   
 \\
\midrule
CSD          & N/A        & N/A              & 0.848 ± 0.189  \\
\midrule
\multirow{3}{*}{voxel-wise}  &      &         & 0.838  ± 0.195$\textcolor{red}{\bm{\star}}$    \\
     & \checkmark       &    & 0.849  ± 0.175$\textcolor{red}{\bm{\star}}$         \\
  & \checkmark      & \checkmark     & 0.879  ± 0.138$\textcolor{red}{\bm{\star}}$    \\

\midrule
\multirow{3}{*}{Patch-wise}     &     &     & 0.842 ± 0.185$\textcolor{red}{\bm{\star}}$    \\
& \checkmark   &       &  0.856 ± 0.173$\textcolor{red}{\bm{\star}}$ \\
  & \checkmark   & \checkmark     & \textbf{0.902 ± 0.128}$\textcolor{red}{\bm{\star}}$     \\

\bottomrule
\end{tabular}
\newline
\label{table:performance2}
\caption*{Table 3: Model/Method degradation is evaluated by assessing the prediction consistency with fewer diffusion direction signals. Wilcoxon signed-rank test is applied as a statistical assessment. The statistical difference ($p < 0.001$) compared with full direction reconstruction from CSD is marked as $\textcolor{red}{\bm{\star}}$. }
\end{table}

\subsection{Downstream task evaluation}
The APOE states are predicted using different brain structure connectome maps. This task is employed as a downstream task to evaluate the performance of different fODF modeling methods. Briefly, we calculated the accuracy, precision, recall, and F1 for each method via 20 bootstraps. Then, the 95\% confidence intervals are calculated for all metrics to demonstrate the variability. As shown in Table~\ref{apoe}, all lower bounds of deep learning metrics are higher than the upper bounds of the CSD method.


\begin{table}[h]
\centering{}
\caption{Confidence interval on metrics of biomarker predictions}
\begin{tabular}{c|ccccccc}
\toprule
  \cellcolor[gray]{0.9}       & \multirow {2}{*}{DL}   & \multicolumn{2}{c}{95\% CI}     &  \multirow {2}{*}{CSD}    & \multicolumn{2}{c}{95\% CI} & \multirow {2}{*}{$p$-value}  \\       
\cline{3-4} \cline{6-7}    
\cellcolor[gray]{0.9} Metrics  &   & \makecell{Lower\\bound} & \makecell{Upper\\bound} &   & \makecell{Lower\\bound} & \makecell{Upper\\bound}  \\
\midrule
\cellcolor[gray]{0.9}Accuracy & 0.568 & [0.560 & 0.577] & 0.526 & [0.517 & 0.535]   & $p<0.001$ \\
\cellcolor[gray]{0.9}Precision & 0.520 & [0.513 & 0.528] & 0.463 & [0.456 & 0.472]   & $p<0.001$ \\
\cellcolor[gray]{0.9}Recall & 0.525 & [0.515 & 0.534] & 0.450 & [0.438 & 0.461]   & $p<0.001$\\
\cellcolor[gray]{0.9}F1 & 0.511 & [0.502 & 0.520] & 0.441 & [0.432 & 0.480]          & $p<0.001$\\
\bottomrule
\end{tabular}
\caption*{Table 4: The Mean and the range of the 95\% confidence interval for the deep learning metrics are reported. The $p$-value of the metrics from both methods are statistically significant.}
\label{apoe}
\end{table}

\section{Discussion}
Ensuring scan-rescan variability is of paramount importance in diffusion signal modeling, particularly when employing a data-driven approach. Data-driven methods, which learn directly from the data, inherently rely on the consistency and reliability of the input data to produce accurate and robust models. As such, the issue of scan-rescan variability comes into sharp focus. In the context of diffusion signal modeling, variations between repeated scans of the same subject can introduce inconsistencies that could significantly affect the performance of the data-driven models. This is because these models are sensitive to the statistical properties of the training data. If the input data are inconsistent due to scan-rescan variability, the learned model may not generalize well, leading to less accurate predictions. Thus, in our study, addressing scan-rescan variability is not just a quality control issue, but a crucial factor that directly impacts the reliability and clinical applicability of data-driven diffusion signal models. A consistent focus on minimizing this variability can result in models that provide more accurate, reliable, and clinically meaningful results.

In our study, we developed a data-driven fODF modeling algorithm to provide robust microstructure estimation for modeling tractography. The proposed method (1) learns the mapping from SH basis 3D DW-MRI signal to a fiber ODF, (2) improves consistency and alleviates the effects that occur between different scanners with a new loss function, (3) increases model robustness in the ’fewer diffusion directions’ scenarios, and (4) empowers better a predicative power in downstream tasks (e.g., APOE states estimation from connectomes).

The improvement from voxel-vise input to patch-vise input is based on the assumption that the model is trained on a diverse patch selection, it sees a wide range of scenarios. And meanwhile, a deep learning model can capture the complex relationships as a deeper network is able to account for the 3D context without being unduly influenced by neighboring voxels in some scenarios. This is the advantage of data-driven method, we do not need to explicitly ensure that neighboring voxels have similar distribution functions to central voxel.

However, there are still several limitations in our approaches. First, one key limitation of our approaches is the cost of computing resources by performing patch-to-center predictions. We need to generate 27 times storage for the 3x3x3 patches for one single DMRI.  Second, to provide more precise microstructure estimation, we need to target our model to multi-tissue multi-shell CSD (MSMT-CSD) which provides different fiber response functions to different tissues. It also leads to a potential future question on how to encode the b-value information into SH representation during the transform of multi-shell DWI and signal ODFS.

In general, our study is a step towards the direct harmonization of the estimated microstructure (FOD) using deep learning and a data-driven scheme when scan-rescan data are available for training. The proposed method is potentially applicable to a wider range of data harmonization problems in neuroimaging.

\section{Conclusion}
In this paper, we propose a novel deep constrained spherical deconvolution method to explicitly reduce the scan-rescan variabilities, so as to model a more reproducible and robust brain microstructure of repeated DW-MRI scans that are acquired from the same patient. From the experimental results, the proposed data-driven framework outperforms the existing benchmarks in fODF estimation. In general, our study is a step towards the direct harmonization of the estimated microstructure (e.g., FOD) using deep learning and a data-driven scheme when scan-rescan data are available for training. The proposed method is potentially applicable to a wider range of data harmonization problems in neuroimaging.

\section*{Disclosures}
The authors of the paper have no conflicts of interest to report. 

\section* {Acknowledgments}
This work was supported by the National Institutes of Health under award numbers
R01EB017230, T32EB001628, and 5T32GM007347, and in part by the National Center for Research Resources and Grant UL1 RR024975-01. This study was also supported by National Science Foundation (1452485, 1660816, and 1750213).
The content is solely the responsibility of the authors and does not necessarily represent the official views of the NIH or NSF. 


\bibliography{main}   
\bibliographystyle{spiejour}   


\vspace{2ex}\noindent\textbf{Tianyuan Yao} is currently a PhD student in Computer Science at Vanderbilt University. He is supervised by Prof. Yuankai Huo at HRLB Lab. He received his bachelor's degree from Shandong University in 2019, with major in Communication Engineering, and MS degree of Computer Science from Vanderbilt University in 2021. His main research interests are medical image analysis, deep learning, and computer vision. He is passionate about their applications in pathology and radiology imaging.

\vspace{1ex}
\noindent Biographies and photographs of the other authors are not available.

\listoffigures
\listoftables

\end{spacing}
\end{sloppypar}
\end{document}